\pdfoutput=1

\documentclass[11pt]{article}

\usepackage[preprint]{acl}

\usepackage{times}
\usepackage{latexsym}

\usepackage[T1]{fontenc}

\usepackage[utf8]{inputenc}
\usepackage{float}
\usepackage{newfloat}
\usepackage{multirow}
\usepackage{makecell}
\usepackage{xcolor}         
\usepackage{colortbl}
\usepackage{dsfont}
\usepackage{amsmath}
\usepackage{enumitem}

\usepackage[normalem]{ulem}
\useunder{\uline}{\ul}{}

\definecolor{myyellow}{rgb}{1,1, 0.6}
\definecolor{myorange}{rgb}{1, 0.8, 0.6}
\definecolor{myred}{rgb}{1, 0.6, 0.6}
\definecolor{second}{HTML}{FFDAB9}
\definecolor{best}{HTML}{FFC1C1}
\definecolor{lightblue}{rgb}{0.8,0.9,1}
\usepackage{multirow}
\usepackage{colortbl}
\usepackage[normalem]{ulem}
\usepackage{bbding}
\useunder{\uline}{\ul}{}
\usepackage{stackrel}
\usepackage{todonotes}
\usepackage{booktabs}
\definecolor{myyellow}{rgb}{1,1, 0.6}
\definecolor{myorange}{rgb}{1, 0.8, 0.6}
\definecolor{myred}{rgb}{1, 0.6, 0.6}
\newcommand*{\method}{HyperLLaVA}

\renewcommand{\todo}[1]{\iffalse #1 \fi{\color{blue} \textbf{[TODO]}}}
\vspace{-5cm}
\title{HyperLLaVA: Dynamic Visual and Language Expert Tuning for Multimodal Large Language Models
}

\author{
    Wenqiao Zhang~\textsuperscript{1 $\spadesuit$}  \And 
    Tianwei Lin~\textsuperscript{2 $\spadesuit$} \And
    Jiang Liu~\textsuperscript{3 $\spadesuit$} \And
    Fangxun Shu~\textsuperscript{ 4} \And
    Haoyuan Li~\textsuperscript{ 1} \AND
    Lei Zhang~\textsuperscript{ 4} \And
    He Wanggui~\textsuperscript{ 4} \And 
    Hao Zhou~\textsuperscript{ 5} \And
    Zheqi Lv~\textsuperscript{ 1} \AND
    Hao Jiang~\textsuperscript{ 4 $\clubsuit$} \And
    Juncheng Li~\textsuperscript{ 1 $\clubsuit$} \And
    Siliang Tang~\textsuperscript{ 1} \And
    Yueting Zhuang~\textsuperscript{  1 $\clubsuit$} \AND
    \vspace{-0.1cm}
\small{\textsuperscript{\rm 1} \textbf{Zhejiang University}  }, 
\small{\textsuperscript{\rm 2} \textbf{ShanghaiTech University}  }, 
\small{\textsuperscript{\rm 3} \textbf{Chongqing University}  }, 
\small{\textsuperscript{\rm 4} \textbf{Alibaba Group}  },
\small{\textsuperscript{\rm 5} \textbf{Harbin Institute of Technology} }
\\
\small{\texttt{\{wenqiaozhang, lihaoyuan, zl.leizhang, zheqilv, junchengli, siliang, yzhuang\}@zju.edu.cn}}\\
\small{\texttt{linjiawei@shanghaitech.edu.cn,jiangliu@stu.cqu.edu.cn,
\{shufangxun.sfx, aoshu.jh\}@alibaba-inc.com}} \\\\
}

\begin{document}
\maketitle
\vspace{+2cm}
\begin{abstract}
Recent advancements indicate that scaling up Multimodal Large Language Models (MLLMs) effectively enhances performance on downstream multimodal tasks. 
The prevailing MLLM paradigm, \emph{e.g.}, LLaVA, transforms visual features into text-like tokens using a \emph{static} vision-language mapper, thereby enabling \emph{static} LLMs to develop the capability to comprehend visual information through visual instruction tuning.
Although promising, the \emph{static} tuning strategy~\footnote{The static tuning refers to the trained model with static parameters.} that shares the same parameters may constrain performance across different downstream multimodal tasks.
In light of this, we introduce HyperLLaVA, which involves adaptive tuning of the projector and LLM parameters, in conjunction with a dynamic visual expert and language expert, respectively.
These experts are derived from HyperNetworks, which generates adaptive parameter shifts through 
visual and language guidance, enabling dynamic projector and LLM modeling in two-stage training.
 Our experiments demonstrate that our solution significantly surpasses LLaVA on existing MLLM benchmarks, including MME, MMBench, SEED-Bench, and LLaVA-Bench. ~\footnote{Our project is available on the link https://github.com/DCDmllm/HyperLLaVA}.
\end{abstract}
\section{Introduction}
\label{sec:Introduction}   
The landscape of Large Language Models (LLMs)~\cite{devlin2018bert,radford2018improving,ouyang2022training} has undergone significant evolution, highlighting their exceptional versatility in managing a wide variety of language-centric applications. To extend the capabilities of LLMs to a wider array of modal inputs, Multimodal Large Language Models (MLLMs) have garnered increasing attention~\cite{ref:CLIP, li2022blip, huang2023language, ref:GPT4, li2023fine}. MLLMs are crucial for the development of flexible, general-purpose assistants, as everyday interactions encompass information from various modalities (\emph{e.g.}, videos, audio, 3D environments, point clouds) in addition to text. 

\begin{figure*}
    \centering
    \includegraphics[width=1\textwidth]{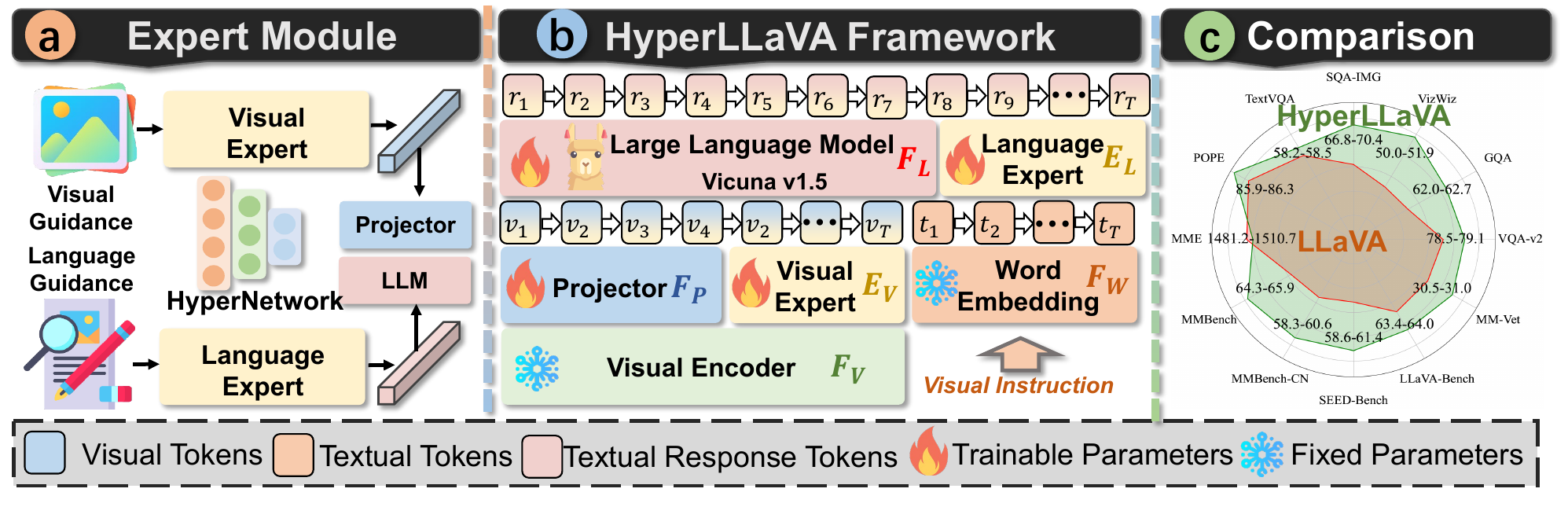}
    \vspace{-0.7cm}
    \caption{(a) is the overview of LLaVA. (b) describes the simplified version of our \method{}. (c) shows that compared to LLaVA, our method achieves superior performance across different MLLM benchmarks.
    }
    \label{fig:introduction}
\end{figure*}

Contemporary MLLMs (\emph{e.g.}, LLaVA~\cite{ref:LLaVA,liu2023improved}) typically adhere to a two-stage training protocol:
(i) \textbf{Vision-Language Alignment}: A static projector is trained by leveraging image-text pairs to synchronize visual features with the language model’s word embedding space.
The projector with static parameters connects the vision and language modalities by translating visual features into visual tokens, enabling the LLM to understand the visual content. The quality of conveyed visual tokens directly influences the MLLM's performance~\cite{zhou2023infmllm}.
(ii) \textbf{Multimodal Insturction Tuning.} Following vision-language alignment, multimodal instruction data are used to refine the LLM, enabling it to respond to users' varied requests involving visual content. This step is crucial for augmenting the capabilities and controllability of MLLM to address different downstream multimodal tasks.

Despite two-stages' critical importance, the projector's structure and LLM tuning strategy have been relatively underexplored, most of the pieces of literature focus on scaling up the pretraining data~\cite{bai2023qwen,ref:InstructBLIP}, instruction-following data~\cite{li2023mimic,zhang2023enhanced,zhao2023svit},  visual encoders~\cite{bai2023qwen} or language models~\cite{lu2023empirical} to facilitate vision-language understanding.  What's more,  further quantitative analyses show that the learned model with static parameters may limit their potential for multi-downstream tasks~\cite{mahabadi2021parameter,zhang2023revisiting}.  
Based on the aforementioned insights, our investigation focuses on the two-stage training process, transitioning from static to dynamic tuning—that is, tuning both the projector and LLM with dynamic parameters to provide flexible design choices that bolster the MLLM's reasoning abilities across diverse multimodal tasks.



In this paper, we propose \method{} (Figure~\ref{fig:introduction}(b)), its dynamic characterization benefits from a carefully designed expert module, which is derived from HyperNetworks~\cite{ref:hypernetworks} to generate the dynamic parameters based on input information. 
Our bootstrapping philosophy is to dynamically generate strongly correlated features according to visual and language guidance, thereby dynamically modeling the projector and LLM layers, respectively. In detail, HyperLLaVA is learned following the two steps: (i) In vision-language alignment, we divide the projector into static layers (original MLP in LLaVA~\cite{liu2023improved}) and dynamic layers (visual expert), where the parameters of static layers are fixed, while the parameters of dynamic
layers are dynamically generated based on visual input. 
The visual expert leverages the Hypernetwork to assist the static projector learn a visual-specific projector that adaptively models the
visual features according to the visual guidance.
By doing so, the projector can deliver the adaptative visual tokens to the language semantic space. 
(2) In the multimodal instruction tuning stage, we equip the LLM with a language expert, modeling dynamic parameters for LLM blocks. We regard the intermediate output of LLM as language guidance that guides the language expert to provide an improved instruction-specific comprehension of the user's request.
By doing so, the MLLM increases the flexibility by instead generating unique parameters for every input, allowing the MLLM to make
use of similarities between samples across datasets and avoid potential interference between samples within the same dataset.
Notably, the proposed language expert serves as a parameter-efficient fine-tuning approach for the MLLMs, yielding a comparable performance according to the original LLaVA. 




In summary, our contributions are three-fold as follows:
\begin{itemize}
\item  We study the under-explored dynamic tuning strategy for MLLMs and introduce HyperLLaVA, leveraging the visual and language-guided dynamic tuning for projector and LLM; \item  The proposed visual and language expert serves as a parameter-efficient method of multitask fine-tuning;
\item  We conducted comprehensive and detailed experiments across multiple MLLM benchmarks. The rich experimental results prove the effectiveness and universality of our proposed method.
\end{itemize}

\section{Related Work}
\label{sec:Related Work} 
\noindent \textbf{Large Language Model.} 
The proliferation of Large Language Models (LLMs) has dramatically reshaped the landscape of natural language processing.
Pioneering models such as encoder-centric model BERT~\cite{devlin2018bert} and decoder-centric model GPT~\cite{radford2018improving} have led this charge, showcasing that enhancements in model size and the expansiveness of training datasets can result in unprecedented improvements in performance. 
Building on the achievements of their predecessors, subsequent models have brought forth substantial innovations that further advance the prowess of LLMs. PaLM~\cite{chowdhery2023palm} highlighted the benefits of increasing model parameters for enhanced language comprehension. Meanwhile, InstructGPT~\cite{ouyang2022training} and ChatGPT utilized fine-tuning and reinforcement learning strategies to refine their performance in conversational interaction~\cite{chen2023sharegpt4v}.
However, the reliance on textual data as the sole source of learning has been a limiting factor, as it constrains the models' ability to engage with the richly interconnected world of multimodal information.\\
\noindent \textbf{Multimodal Large Language Model.} In recent years, the development of deep learning has brought prosperity to the field of multimodal intelligence~\cite{baltruvsaitis2018multimodal,li2023variational,zhang2022boostmis,zhang2019frame,zhang2022magic}.
Multimodal Large Language Models (MLLMs) leverage the power of LLMs, mitigating extra computational cost and enhancing the efficacy of multimodal pre-training~\cite{ref:MM-LLMs}, to bridge the gap between textual and multimodal data(\emph{e.g.}, images, videos, and audios). A prominent attempt is CLIP~\cite{ref:CLIP}, demonstrating the alignment of visual and textual modalities via contrastive learning across a broad dataset of image-text pairs.~\cite{li2022blip} and ~\cite{li2023blip} follow this trend, proposing BLIP and BLIP-2 improved upon CLIP, and gain remarkable performance in basic visual tasks. 
Flamingo~\cite{ref:flamingo} led the way in merging vision and language models by utilizing vast amounts of intertwined image-text dataset, revealing unparalleled zero-shot capabilities in processing image-text content within conversational contexts for the first time. 
LLaVA~\cite{ref:LLaVA} distinctively incorporates short captions annotated by humans and bounding boxes into the GPT4 language model. In the realm of audio processing, there are also some brilliant works, such as SpeechT5~\cite{ref:SpeechT5}, MMS~\cite{ref:MMS}, PandaGPT~\cite{ref:PandaGPT}, etc.\\
\noindent \textbf{Hypernetworks.}
The original HyperNetwork~\cite{ref:hypernetworks} is designed to reduce the number of parameters, \emph{i.e}, a small neural network generates parameters for another big neural network, thereby obtaining the model compression for different tasks. Subsequently, HyperNetwork is developed to various domain tasks, including few-shot learning~\cite{ref:hypernetwork_one_shot},  graph modeling~\cite{ref:hypernetwork_graph}, domain adaptation~\cite{zhang2023revisiting}, device-cloud collaboration~\cite{ref:hypernetwork_duet,hypernetwork_ideal}, etc.

\section{Methodology}
This section describes the proposed MLLM framework HyperLLaVA. We shall present each module and its training strategy.
\label{sec:method}

\subsection{Problem Formulation}
The primary goal is to effectively leverage the capabilities of both the pre-trained LLM and visual model. 
The network architecture is illustrated in Figure 2. Given an RGB image $x \in R^{H \times W \times 3}$, where $H$ and $W$ are the origin resolution. 
The vision encoder processes input images to obtain a visual token sequence $\mathcal{V}=[v_1, v_2, \cdots, v_{N_v}]$, where ${N_v}$ represents the sequence length of text tokens. 
Subsequently, we concatenate the visual tokens and text tokens $\mathcal{T}=[t_1, t_2, \cdots, t_{N_t}]$, together and feed them into a LLM $\mathcal{M}_llm$, then generate the language response $\mathcal{R}=[r_1, r_2, \cdots, t_{N_r}]$,  where $N_t$ and $N_r$ indicate the length of text tokens and textual response.
In general, MLLM model $\mathcal{M}(\cdot)$ consists of two functions as below:
\begin{equation}
\begin{aligned}
\underbrace{\mathcal{M}(\cdot)}_{\rm{MLLM}}: \underbrace{\mathcal{M}_p((\mathcal{T}|\mathcal{V});\Theta_{p})}_{\rm{Projector}}  \rightarrow
\underbrace{\mathcal{M}_l((\mathcal{R}|\mathcal{V},\mathcal{T});\Theta_{{l}})}_{\rm{LLM}}\,,
 \label{adv}
\end{aligned}
\end{equation}
where $\mathcal{M}_{p}(\cdot;\Theta_{p})$ is the  projector and $\mathcal{M}_t(\cdot;\Theta_{l})$ LLM tuning with multi-modal instructions with parameters $\Theta_{p}$ and $\Theta_{l}$, respectively.

\subsection{Preliminaries}
\label{sec:preliminary}
\noindent\textbf{LLaVA.} 
\label{sec:LLaVA} 
\emph{LLaVA}~\cite{ref:LLaVA} is trained following two steps: (i) First,
a two-layer MLP is employed as vision-language projector $\mathcal{M}_p(\cdot)$ to convert visual features into
visual tokens $V$, which have the same dimensionality as the word embedding space
in the language model. (ii) Then LLaVA performs instruction-tuning with visual tokens $V$ and language tokens $T$ for the LLM (Llama) $\mathcal{M}_l(\cdot)$, generating  response tokens $\mathcal{R}$ by optimizing its  auto-regressive training objective.

\noindent\textbf{HyperNetwork.} 
\label{sec:hyp} \textit{Hypernetwork}~\cite{ha2016hypernetworks} is a neural network that generates the weights for another neural network.
Specifically, HyperNetwork treats the parameters of the multi-layer perception (MLP) as a matrix  $K^{(n)} \in {R}^{N_{in} \times N_{out}}$, where $N_{in}$ and $N_{out}$ represent the number of input and output neurons of the $n^{th}$ layer of MLP, respectively. $N_{in}$ and $N_{out}$ portray the structure of the MLP layers together. The generation process of $K^{(n)}$ can be regarded as a matrix factorization:
 \begin{align}
    K^{(n)} = \xi(z^{(n)};\Theta_p), \forall n=  1, \cdots, N_l\,.
\end{align}
In the \emph{training procedure}, $z^{(n)}$ and $\xi(\cdot)$ are randomly initialized. 
The gradients are backpropagated to $z^{(n)}$ and $\xi(\cdot)$, which can help to update them. $z^{(n)}$ and $\xi(\cdot)$ will be saved instead of $K^{(n)}$.

\subsection{Vision-language Guided Expert Module}
\label{sec:expert}
Original LLaVA's projector and LLM are trained with static parameters. We argue that the static tuning paradigm may limit the flexible visual token delivery and appropriate expression for different downstream multi-modal tasks. Thus, we propose to equip the original's LLaVA projector and LLM with a visual expert $\mathcal{E}_V$ and a language expert $\mathcal{E}_L$: (i) the visual expert adaptively fits the projector's output according to the specific visual guidance (\emph{e.g}, visual features); (ii) the language expert dynamically modeling the posterior blocks of LLM through anterior LLM's block output. 

\begin{figure*}
    \centering
    \includegraphics[width=1\textwidth]{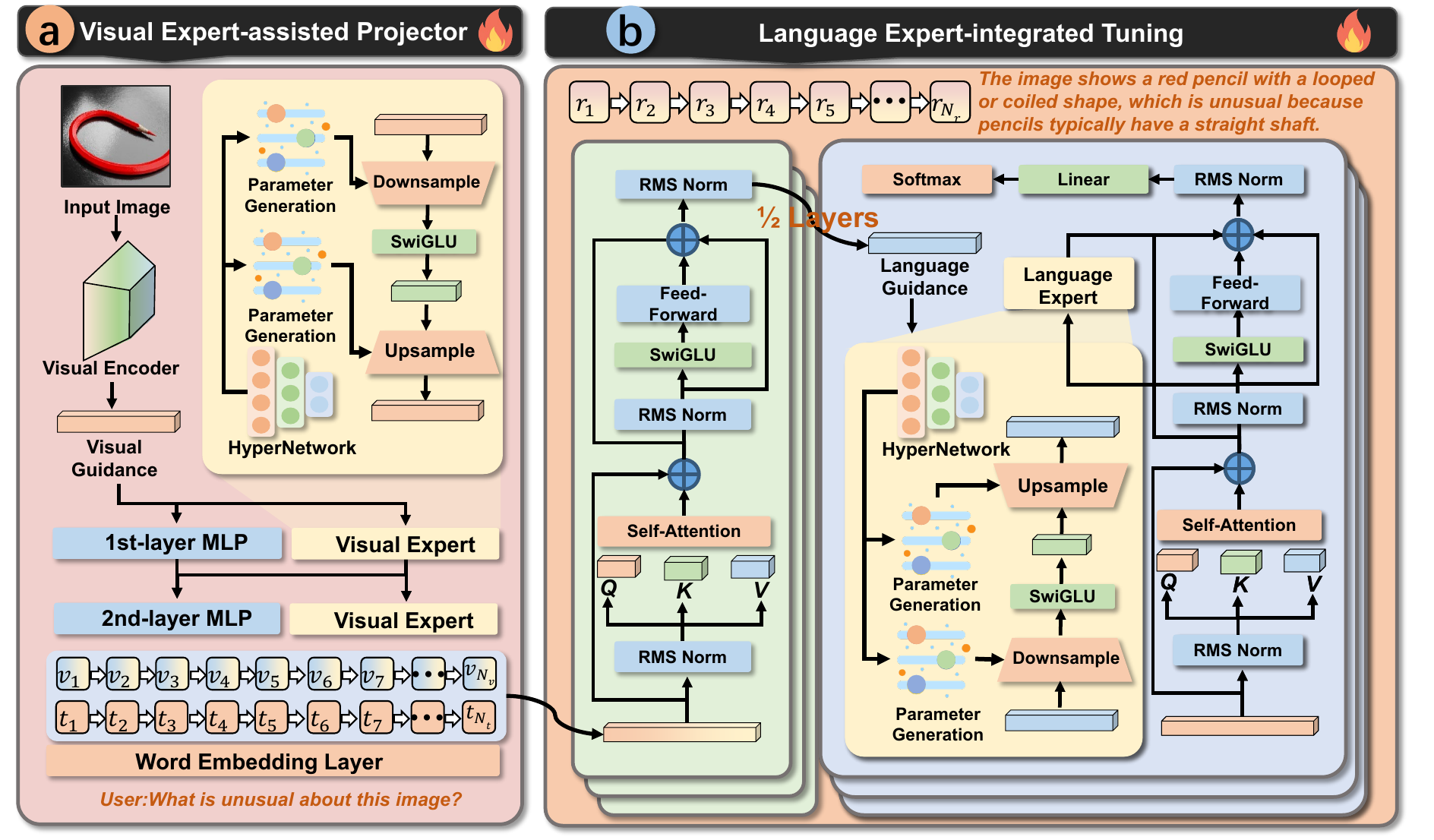}
    \caption{ Overview of proposed \method{}. (a) describes how the proposed visual expert assists the static projector that dynamically converts the image features to adaptive visual tokens, yielding an augmented visual expression for subsequent instruction tuning. (b) is the proposed language expert-integrated tuning, which uses the output of the intermediate layer as language guidance to generate dynamic instruction-specific feature, increasing the flexibility for processing different multimodal tasks.
    }
    \label{fig:framework}
\end{figure*}
The expert module is derived from Hypernetorks, which is a neural network that generates its parameters for another neural network. As HyperNetwork dynamically generates a network conditioned on the input embeddings, \emph{i.e.}, the ``dynamic characterization'' can be modeled by HyperNetwork.
However, directly utilizing the HyperNetwork may not satisfactorily model dynamic learning for two key reasons: 

\begin{itemize}
\item \textbf{Weak Correlation.} The original HyperNetwork learns the latent vector to generate another model's parameters. This lacks a strong correlation between parameter generation and input guidance.

\item \textbf{Unstable Optimization.} Using HyperNetwork generate the parameters of the projector or LLM block is large ($D_x \times N_{in} \times N_{out}$), \emph{i.e.}, it is hard to optimize the such the numerous parameters, the optimization process is intuitively unstable. 
\end{itemize}

To this end, we carefully tailor the HyperNetwork with the following adjustments: 

\noindent\textbf{Input Prior Guidance.} We first propose to model the dynamic layers by replacing the learned latent vector $z$ with specific input.  
Specifically, given the feature $f_{x^{(i)}}$ extracted from backbone of sample $x^{(i)}$, we first develop a layer-specific encoder $E^{n}(\cdot)$ that encode the $f_{x^{(i)}}$ as $\boldsymbol{e}^{(n)}$. This vector 
 represents the $n^{th}$ layer parameters.
 \begin{align}
\label{eq:kernal_generation}
\begin{gathered}
 \boldsymbol{e}^{(n)} = {E}^{n}(f_{x^{(i)}}), \forall n= 1, \cdots, N_l\,,
\end{gathered}
\end{align}
  where $N_l$ is the number of the modeled layers. 

  Then the HyperNetwork is used to convert the embedding $\textbf{e}^{(n)}$ into parameters, \emph{i.e.}, we input $\textbf{e}^{(n)}$ into the following two MLP layers to generate parameters of dynamic layers. 
\begin{align}
\label{eq:kernal_generation_detail}
\begin{gathered}
  \mathcal{\boldsymbol{w}}^{(n)} = (W_1\boldsymbol{e}^{(n)} + B_1)W_2 + B_2, \\
  K^{(n)} = \mathcal{\boldsymbol{w}}^{(n)} + \mathcal{\boldsymbol{b}}^{(n)},
\end{gathered}
\end{align}
where $K^{(n)}$ denotes the $n^{th}$ layer parameters of dynamic layers. Two MLP layers's weight are denoted by $W_1$ and $W_2$, respectively. $\mathcal{\boldsymbol{b}}^{(n)}$, $B_1$ and $B_2$ are the biases.

\noindent\textbf{HyperNetwork-aware Adapter.}  Adapters are sub-networks with small parameters that are inserted after every attention and feed-forward layer in a model~\cite{houlsby2019parameter}. The original adapter is a parameter-efficient learning approach that learns downstream tasks by updating only a small number of parameters. The adapters consist of a pair of downsampling and upsampling layers, and a residual connection. We found that using downsampling and upsampling strategies, the HyperNetwork-generated parameters can be substantially reduced. 

Given the visual guidance $x_V$ and language guidance $x_L$, the vision-language guided expert can be defined as:
 \begin{align}
\label{eq:adapter}
\begin{gathered}
\mathcal{E}_M(x_M)=W_M^u({\rm SwiGLU}(W_M^d(x_M)))\\  W_M^u,W_M^d=\mathcal{H}_M(x_M), {\rm where} M\in {V,L} 
\end{gathered}
\end{align}
where $M$ indicate the modality, $W_M^u,W_M^d$ respectively denote the weights for upsampling and downsampling. SwiGLU~\cite{ramachandran2017searching} is the activation function, Gaussian Error Linear Unit. $\mathcal{H}_M$ is the HyperNetwork.
 
\subsection{Visual Expert-Assisted Projector}
\label{sec:projector}


In this stage, our objective is to adapt the image tokens to LLM, allowing the LLM to comprehend the instances in the images. As shown in Figure~\ref{fig:framework}, we divide the projector as static layers and dynamic layers. Following LLaVA1.5~\cite{liu2023improved}, we employ an two-layer MLP as the static layers. To empower the projector's expression, we develop a visual expert that learning the projector shift to model the dynamic text tokens. Specifically, given the visual feature $f_V$ extracted from visual encoder, the visual expert will adaptively convert $f_V$ to dynamic visual embeddings. We show three alternatives for the dynamic vision-language alignment, the visual tokens $V$ can be calculated as: 
\begin{equation}  V=
\begin{cases} 
\underbrace{\mathcal{L}_{2}(\mathcal{L}_{1}(f_V)+\mathcal{E}_{V_1}(f_V))}_{\mbox{Use 1st Visual Expert}} \\
\underbrace{\mathcal{L}_{2}(\mathcal{L}_{1}(f_V))+\mathcal{E}_{V_2}(\mathcal{L}_{1}(f_V))}_{\mbox{Use 2nd Visual Expert}}\\
\underbrace{\mathcal{L}_{2}(\mathcal{L}_{1}(f_V)+\mathcal{E}_{V_1}(f_V))+\mathcal{E}_{V_2}(\mathcal{L}_{1}(f_V))}_{\mbox{Use 1st\&2nd Visual Expert}}
\end{cases} 
\label{computeA}
\end{equation}
where $\mathcal{L}_1$ and $\mathcal{L}_2$ denotes two MLPs, $\mathcal{E}_{V_1}$ and $\mathcal{E}_{V_2}$ are the visual expert for first and second MLPs. We give a details comparison in the Sec. experiments.

\label{tab:main_results}

\begin{table*}
        \centering
        \captionsetup{font={small,stretch=1.25}, labelfont={bf}}
        \caption{\textbf{Comparison with SoTA methods on 12 benchmarks.} Res, PT, IT indicate input image resolution, the number of samples in the pretraining and instruction tuning stage,
    respectively. We color each row as the \colorbox{myred}{\textbf{best}} and \colorbox{myorange}{\textbf{second best}}. Improv. $\uparrow$ indicates performance improvement compared with LLaVA. }
            \resizebox{\textwidth}{!}{%
            \renewcommand{\arraystretch}{1.25}
            \begin{tabular}{ccccc||ccccc||ccccccc}
            \Xhline{1.5pt}
            \multirow{2}{*}{\textbf{\Large Method}} & \multirow{2}{*}{\textbf{\Large LLM}} & \multirow{2}{*}{\textbf{\Large Res.}} & \multirow{2}{*}{\textbf{\Large PT}} & \multirow{2}{*}{\textbf{\Large IT}} & \multicolumn{5}{c||}{\texttt{VQA Datasets}} & \multicolumn{7}{c}{\texttt{Benchmark Toolkits}} \\
            \Xcline{6-10}{0.2pt} \Xcline{11-17}{0.2pt}
            & & & & & \textbf{VQA\textsuperscript{v2}} & \textbf{GQA} & \textbf{VizWiz} & \textbf{SQA\textsuperscript{I}} & \textbf{VQA\textsuperscript{T}} & \textbf{POPE} & \textbf{MME} & \textbf{MMB} & \textbf{MMB\textsuperscript{CN}} & \textbf{SEED} & \textbf{LLaVA\textsuperscript{W}} & \textbf{MM-Vet} \\

            \Xhline{1.5pt}
            BLIP-2~\cite{li2023blip} & Vicuna-13B & 224 & 129M & - & 41.0 & 41 & 19.6 & 61 & 42.5 & 85.3 & 1293.8 & - & - & 46.4 & 38.1 & 22.4 \\
            InstructBLIP~\cite{ref:InstructBLIP} & Vicuna-7B & 224 & 129M & 1.2M & - & 49.2 & 34.5 & 60.5 & 50.1 & - & - & 36 & 23.7 & 53.4 & 60.9 & 26.2 \\
            InstructBLIP~\cite{ref:InstructBLIP} & Vicuna-13B & 224 & 129M & 1.2M & - & 49.5 & 33.4 & 63.1 & 50.7 & 78.9 & 1212.8 & - & - & 58.2 & - & 25.6 \\
            Shikra~\cite{ref:Shirka} & Vicuna-13B & 224 & 600K & 5.5M & 77.4 & - & - & - & - & - & 58.8 & - & - & - & - & - \\
            IDEFICS-9B~\cite{laurencon2023obelics} & LLama-7B & 224 & 353M & 1M & 50.9 & 38.4 & 35.5 & - & 25.9 & - & - & 48.2 & 25.2 & - & - & - \\
            IDEFICS-80B~\cite{laurencon2023obelics} & LLama-65B & 224 & 353M & 1M & 60.0 & 45.2 & 36.0 & - & 30.9 & - & - & 54.5 & 38.1 & - & - & - \\
            Qwen-VL~\cite{ref:Qwen-VL} & Qwen-7B & 448 & 1.4B & 50M & 78.8 & 59.3 & 35.2 & 67.1 & \cellcolor{myred}{63.8} & - & - & 38.2 & 7.4 & 56.3 & - & - \\
            Qwen-VL-Chat~\cite{ref:Qwen-VL} & Qwen-7B & 448 & 1.4B & 50M & 78.2 & 57.5 & 38.9 & 68.2 & \cellcolor{myorange}{61.5} & - & 1487.5 & 60.6 & 56.7 & 58.2 & - & - \\
            \Xhline{1pt}
            \method{} w/o $\mathcal{E}_V$ (Ours)  & Vicuna-7B & 336 & 558K & 665K & \cellcolor{myorange}{79.0} & \cellcolor{myorange}{62.5} & 50.3 & \cellcolor{myorange}{70.4} & 58.1 & \cellcolor{myorange}{85.9} & 1486.0 & \cellcolor{myorange}{65.9} & \cellcolor{myorange}{59.7} & \cellcolor{myorange}{61.0} & \cellcolor{myorange}{63.7} & \cellcolor{myred}{32.8} \\
             \method{} w/o $\mathcal{E}_L$ (Ours) & Vicuna-7B & 336 & 558K & 665K & 78.8 & 61.9 & \cellcolor{myred}{52.1} & \cellcolor{myred}{70.7} & 57.5 & 85.6 & \cellcolor{myorange}{1492.0} & \cellcolor{myred}{66.7} & 58.6 & 60.8 & 62.6 & 30.9 \\
            \Xhline{1pt}
             LLaVA-1.5~\cite{liu2023improved} & Vicuna-7B & 336 & 558K & 665K & 78.5 & 62.0 & 50.0 & 66.8 & 58.2 & \cellcolor{myorange}{85.9} & \cellcolor{myred}{1510.7} & 64.3 & 58.3 & 58.6 & 63.4 & 30.5 \\
            \rowcolor{gray!40} \textbf{\method{}} (Ours)  & {Vicuna-7B} & {336} & {558K} & {665K} & \cellcolor{myred}{\textbf{79.1}} & \cellcolor{myred}{\textbf{62.7}} & \cellcolor{myorange}{\textbf{51.9}} & \cellcolor{myorange}{\textbf{70.4}} & \textbf{58.5} & \cellcolor{myred}{\textbf{86.3}} & \textbf{1481.2} & \cellcolor{myorange}{\textbf{65.9}} & \cellcolor{myred}{\textbf{60.6}} & \cellcolor{myred}{\textbf{61.4}} & \cellcolor{myred}{\textbf{64.0}} & \cellcolor{myorange}{\textbf{31.0}} \\

            \rowcolor{lightblue} \textbf{Improv. $ \uparrow$} & \textbf{-} & \textbf{-} & \textbf{-} & \textbf{-} & \textbf{0.6} & \textbf{+0.7} & \textbf{+1.9} & \textbf{+3.6} & \textbf{+0.3} & \textbf{+0.4} & \textbf{-} & \textbf{+1.6} & \textbf{+2.3} & \textbf{+2.8} & \textbf{+0.6} & \textbf{+0.5} \\

            \Xhline{1.5pt}
            \vspace{-0.5cm}
            
            \end{tabular}
            }    
        \label{tab:main_results}
    \end{table*}

    

    \begin{table}
        \begin{center}

     \captionsetup{font={small,stretch=1.25}, labelfont={bf}}
     \caption{\textbf{Three Alternatives for Dynamic Vision-language Alignment.} $\mathcal{E}_{V_1}$ and $\mathcal{E}_{V_2}$ denote visual expert for first and second MLP layer.}
     \renewcommand{\arraystretch}{1.25}
       \resizebox{0.48\textwidth}{!}{
      \begin{tabular}{c||ccc||cc}
        \Xhline{1.5pt}
        \multirow{2}{*}{\textbf{Methods}}  & \multicolumn{3}{c||}{\texttt{VQA Datasets}} & \multicolumn{2}{c}{\texttt{Benchmark Toolkits}} \\
        \Xcline{2-4}{0.5pt} \Xcline{5-6}{0.5pt}
        & \textbf{GQA} & \textbf{SQA-I} & \textbf{VQA-T} & \textbf{POPE}  & \textbf{MME}\\
        \Xhline{1.5pt}
        w/o $\mathcal{E}_V$& 62.5 & 70.4 & 58.1 & 85.9 & 1486.0 \\
       $\mathcal{E}_{V_2}$& 62.0 & 69.8 & 58.0 & 86.4 & 1442.6 \\
       $\mathcal{E}_{V_1}\&\mathcal{E}_{V_2}$& 60.1 & 69.5 & 54.4 & 86.1 & 1449.8 \\
        \Xhline{1pt}
        \rowcolor{gray!40}\textbf{$\boldsymbol{\mathcal{E}_{V_1}}$}& \textbf{62.7} & \textbf{70.4} & \textbf{58.5} & \textbf{86.3} & \textbf{1481.2} \\

       \Xhline{1.5pt}
      \end{tabular}
          \label{tab:visual_expert}}
        \end{center}
        \vspace{-0.5cm}

    \end{table}
Such visual experts learn the projector shift to model the dynamic text tokens, and thus empower the projector's expression for downstream tasks.
\subsection{Language Expert-integrated Tuning}
In this stage, LLM is adjusted to become an LVLM with multi-modal understanding. We use more complex instructions, including tasks such as image logical reasoning and text recognition, which require the model to have a stronger multi-modal understanding. Different
Previous studies have shown that features provided by the intermediate layer may suffice to preliminarily understand the given input samples~\cite{xin2020deebert}and can serve as guidance hints to improve
training~\cite{romero2014fitnets}. Thus, generating guidance in the intermediate LLM layer allows the model to form a preliminary understanding of the given instruction.  Therefore, we regard the output of the intermediate LLM layer as language guidance that generates adaptive instruction-specific features that enhance the generation accuracy. As shown in Figure~\ref{fig:framework}, given the language guidance $f_L$, the adapter's parameters $\{W_L^u, W_L^d\}$ are generated by $\mathcal{H}_L(f_L)$. By doing so, the instruction-specific features can be calculated as below:
\begin{align}
\label{eq:language_expert}
\begin{gathered}
\hat{x}_L=\mathcal{E}_L(x_L)+x_L+{\rm FFN}({\rm SwiGLU}(x_l))
\end{gathered}
\end{align}
where $x_L$ is the features generated from RMS normalization and self-attention in LLM's block.


\section{Experiments}
\label{sec:Experiments}

We verify \method{}'s effectiveness on multiple datasets and then discuss \method{}’s properties with controlled studies. 
\subsection{Dataset and Setting}
\label{sec:setting}
\noindent\textbf{Benchmark Datasets.} We evaluate our proposed \method{} on five VQA datasets:
VQA-v2~\cite{ref:VQA}; GQA~\cite{ref:GQA}; VizWiz~\cite{ref:VizWiz}; SQA\textsuperscript{I}: ScienceQA-IMG~\cite{ref:SQA}; VQA\textsuperscript{T}~\cite{ref:VQAT}: TextVQA and seven Benchmark Toolkits:
POPE~\cite{ref:POPE}; MME~\cite{ref:MME}; MMB: MMBench~\cite{ref:MMB}; MMB\textsuperscript{CN}: MMBench-Chinese~\cite{ref:MMB}; SEED: SEED-Bench~\cite{ref:SEED}; LLaVA\textsuperscript{W}: LLaVA-Bench(In-the-Wild)~\cite{ref:LLaVA}; MM-Vet~\cite{ref:MM-Vet}.

\noindent\textbf{Implementation Details.}  
The model was trained on an 8-A100 machine in one day. The implementation details refer to the Appendix.
In the training of the \method{}, we utilize the ADAMW~\cite{loshchilov2017decoupled} optimizer, adapting hyperparameters to cater to the specific requirements of each phase. For the feature alignment stage, parameters are set as \(B=32\), \(Lr=0.001\), while for visual instruction tuning stage, we adjust the parameters to \(B=16\), \(Lr=0.00002\). The configuration for the ADAMW optimizer incorporates the following settings: \(\boldsymbol{\beta} = (0.9, 0.999)\), \(\varepsilon = 1 \times 10^{-8}\), and $W_d$ = 0.0, ensuring a bespoke optimization strategy that effectively addresses the unique demands of each training phase.

Besides,
We train our model following the same training process as LLaVA-1.5. The process includes two stages: (1) feature alignment stage: use 558K subset of the LAION-CC-SBU dataset to connect a frozen pretrained vision encoder to a frozen LLM; (2) visual instruction tuning stage: use 150K GPT-generated multimodal instruction-following data, plus around 515K VQA data from academic-oriented tasks, to teach the model to follow multimodal instructions.

\noindent\textbf{Comparison of Methods.} For quantifying the efficacy of the proposed framework, we compare \method{} with previous SOTA approaches. We choose BLIP-2\cite{li2023blip}, InstructBLIP\cite{ref:InstructBLIP} based on Vicuna-7B, InstructBLIP\cite{ref:InstructBLIP} based on Vicuna-13B, Shikra~\cite{ref:Shirka}, IDEFICS-9B\cite{laurencon2023obelics}, IDEFICS-80B~\cite{laurencon2023obelics}, Qwen-VL~\cite{ref:Qwen-VL}, Qwen-VL-Chat~\cite{ref:Qwen-VL} and LLaVA-1.5~\cite{liu2023improved}.  More details of baselines are in the Appendix.
    


    \begin{table}
    \centering
     \captionsetup{font={small,stretch=1.25}, labelfont={bf}}
     \caption{\textbf{Analysis of Language Expert Integration for Different LLM Layers.}}
     \renewcommand{\arraystretch}{1.25}
       \resizebox{0.48\textwidth}{!}{
      \begin{tabular}{c||ccc||cc}
       \Xhline{1.5pt}
        \multirow{2}{*}{\textbf{Method}}  & \multicolumn{3}{c||}{\texttt{VQA Datasets}} & \multicolumn{2}{c}{\texttt{Benchmark Toolkits}} \\
        \Xcline{2-4}{0.5pt} \Xcline{5-6}{0.5pt}
        & \textbf{GQA} & \textbf{SQA-I} & \textbf{VQA-T} & \textbf{POPE}  & \textbf{MME}\\
       \Xhline{1.5pt}
        w/o $\mathcal{E}_L$& 61.9 & 70.7 & 57.5 & 85.6 & 1492.0 \\
       Anterior 16 Blocks& 62.5 & 69.4 & 58.5 & 85.9 & 1481.4 \\
       All 32 Blocks & 62.7 & 69.5 & 58.6 & 86.0 & 1460.3 \\
       \Xhline{1pt}
      \rowcolor{gray!40} Posterior 16 Blocks& \textbf{62.7} & \textbf{70.4} & \textbf{58.5} & \textbf{86.3} & \textbf{1481.2} \\
        \Xhline{1.5pt}
      \end{tabular}}
    \label{tab:language_expert}
    \end{table}
    
\begin{table*}
        \centering
        \captionsetup{font={small,stretch=1.25}, labelfont={bf}}
        \caption{\textbf{Zero-shot object hallucination evaluation results on POPE dataset.} "Yes" indicates the proportion of positive responses to the given question. }
            \resizebox{\textwidth}{!}{%
            \renewcommand{\arraystretch}{1.25}
            \begin{tabular}{c||cc||ccc||ccc||ccc}
            \Xhline{1.5pt}
            \multirow{2}{*}{\textbf{ Method}} & \multirow{2}{*}{\textbf{ LLM}} & \multirow{2}{*}{\textbf{ Activated}}  & \multicolumn{3}{c||}{\textbf{Adersarial}} & \multicolumn{3}{c||}{\textbf{Popular}} & \multicolumn{3}{c}{\textbf{Random}} \\
            \Xcline{4-6}{0.2pt} \Xcline{7-9}{0.2pt} \Xcline{10-12}{0.2pt}
            & & & Acc & F1-Score & Yes & Acc & F1-Score & Yes & Acc & F1-Score & Yes \\

            \Xhline{1.5pt}

            mPLUG-Owl~\cite{ref:mPLUG-Owl} & LLaMA-7B & 6.7B & 82.4 & 81.6 & 45.2 & 85.5 & 84.3 & 42.1 & 86.3 & 85.3 & 42.3 \\
            MM-GPT~\cite{ref:MM-GPT} & LLaMA-7B & 6.7B & 50.0 & 66.7 & 100.0 & 50.0 & 66.7 & 100.0 & 50.0 & 66.7 & 100.0 \\
            LLaVA-1.5~\cite{liu2023improved} & Vicuna-7B & 7B & 85.1 & 84.2 & 44.0 & 87.2 & 86.1 & 41.9 & 50.3 & 45.9 & 41.9 \\
            \Xhline{1pt}
            \rowcolor{gray!40} \textbf{\method{}} & {Vicuna-7B} & {7B} & \textbf{85.6} & \textbf{84.7} & \textbf{44.1} & \textbf{87.3} & \textbf{86.2} & \textbf{42.4} & \textbf{50.7} & \textbf{46.5} & \textbf{42.1} \\

            \Xhline{1.5pt}
            \end{tabular}
            }    
         \vspace{-0.5cm}
            
        \label{tab:object_hallucination}
    \end{table*}
\begin{table}
    \begin{center}
 \captionsetup{font={small,stretch=1.25}, labelfont={bf}}
 \caption{\textbf{Deep analysis of expert structure.}}
 \renewcommand{\arraystretch}{1.25}
   \resizebox{0.48\textwidth}{!}{
  \begin{tabular}{c||c c c|| c c}
   \Xhline{1.5pt}
    \multirow{2}{*}{\textbf{Method}}  & \multicolumn{3}{c||}{\texttt{VQA Datasets}} & \multicolumn{2}{c}{\texttt{Benchmark Toolkits}} \\
    \Xcline{2-4}{0.5pt} \Xcline{5-6}{0.5pt}
    & \textbf{GQA} & \textbf{SQA-I} & \textbf{VQA-T} & \textbf{POPE}  & \textbf{MME}\\
    \Xhline{1.5pt}
   Adapter~\cite{houlsby2019parameter}& 57.7& 69.4& 53.5& 83.5& 1371.8\\
   Hypernetwork+MLP & 60.2& 68.8& 52.9& 84.6& 1460.7\\
   Hypernetwork+Adapter~\cite{mahabadi2021parameter} & 62.1 & 69.9 & 57.0 & 85.4 & 1494.4\\
     \Xhline{1pt}
    \rowcolor{gray!40} \textbf{Ours}& \textbf{62.7}& \textbf{70.4}& \textbf{58.5}& \textbf{86.3}& \textbf{1481.2}\\

    \Xhline{1.5pt}
  \end{tabular}
  \label{tab:expert}}
    \end{center}
  \vspace{-0.5cm}

\end{table}

\subsection{Overall Performance}
We benchmark \method{} on a wide range of academic VQA benchmarks and recent benchmarks specifically proposed for instruction-following LMMs, totaling 12 benchmarks.
Table~\ref{tab:main_results} summarizes the quantitative results of our framework and baselines on five \texttt{VQA datasets} and five \texttt{Benchmark Toolkits}. We make the following observations: 
1) In general, irrespective of the different scenarios, compared to LLaVA, \method{} achieves the best performance on almost all the multimodal scenarios across both datasets (Except for the MME benchmark), which
strongly demonstrates the generalizability of the proposed \method{}.
2) \method{} (both 7B and 13B) outperforms bigger MLLMs with billions of trainable parameters for cross-modal connection (\emph{e.g.}, 80B IDEFICS~\cite{laurencon2023obelics}). This further indicates the effectiveness of the proposed MLLM structure. 3) Compared with the original LLaVA, we show that \method{} achieves the best performance across 11 out of 12 benchmarks. Such results benefit from the carefully designed lightweight visual and language expert, which empowers the static projector and LLM to facilitate different multimodal tasks.

\subsection{Ablation Study}
\noindent\textbf{Effectiveness of Each Component.} 
\label{sec:aba}
Table~\ref{tab:main_results} also illustrate the effectiveness of each component, \emph{i.e.}, visual expert $\mathcal{E}_V$ and language expert $\mathcal{E}_L$. Comparing \method{} and \method{}(-$\mathcal{E}_V$) (Row 11 \emph{v.s} Row 13), the $\mathcal{E}_V$ contributes 2.61\% improvement on mean accuracy. 
Meanwhile, Row 11 indicates that it suffers from 0.94\%, a noticeable performance degradation without the $\mathcal{E}_L$. 
To sum up, we can observe that the improvement of using each module alone is distinguishable. Combining all the
components, our \method{} exhibits steady improvement over the baselines.

\subsection{In-Depth Analysis}
We validate the effectiveness of the proposed two modules through the experiments on GQA, SQA-I, VQA-T, POPE and MME benchmarks.

\begin{table}
    \begin{center}
 \captionsetup{font={small,stretch=1.25}, labelfont={bf}}
 \caption{\textbf{Comparsion of parameter-efficient learning.}}
 \renewcommand{\arraystretch}{1.25}
   \resizebox{0.48\textwidth}{!}{
  \begin{tabular}{c||c c c|| c c}
   \Xhline{1.5pt}
    \multirow{2}{*}{\textbf{Method}}  & \multicolumn{3}{c||}{\texttt{VQA Datasets}} & \multicolumn{2}{c}{\texttt{Benchmark Toolkits}} \\
    \Xcline{2-4}{0.5pt} \Xcline{5-6}{0.5pt}
    & \textbf{GQA} & \textbf{SQA-I} & \textbf{VQA-T} & \textbf{POPE}  & \textbf{MME}\\
    \Xhline{1.5pt}
    LoRa~\cite{ref:lora} & 63.0& 68.4& 58.2& 86.4& 1496.9\\
   Adapter~\cite{houlsby2019parameter}& 42.6& 61.2& 41.0& 81.2& 874.4\\
Hypernetwork+Adapter~\cite{mahabadi2021parameter}  & 49.0& 63.6& 48.3& 84.6& 1140.0\\
    \Xhline{1pt}
    \rowcolor{gray!40} Language Expert & \textbf{62.5}& \textbf{70.4}& \textbf{58.1}& \textbf{85.9}& \textbf{1486.0}\\

    \Xhline{1.5pt}
  \end{tabular}}
    \end{center}
    \vspace{-0.5cm}
\label{tab:effecient}
\end{table}
\noindent\textbf{Three Alternatives for Vision-language Alignment.} To build
insights on the visual expert-assisted projector in \method{}, we perform
an in-depth analysis of three alternatives for dynamic vision-language alignment.
Table~\ref{tab:visual_expert} exhibits the three results. According to our observation, using one visual expert to access the dynamic projection yields the best results. Besides, the other two plans also obtained comparable results, indicating the effectiveness of dynamic projection.

\noindent\textbf{Analysis of Language Expert Integration for Different Blocks.} To deeply analyze the effectiveness of language experts, we study the language expert integration for different blocks in Table~\ref{tab:language_expert}, including anterior 16 blocks (before 1/2 LLM layers), all 32 blocks (all LLM layers) and posterior 16 blocks (after 1/2 LMM layers). Generally speaking, leveraging the language expert integration for the posterior 16 blocks obtained almost the best performance. Besides, 
Row 2 and Row 3 utilize the initial language input as language guidance, obtaining suboptimal results compared with language expert integration for the posterior 16 blocks. Our intuition is that the language guidance might not have gathered sufficient contextual information for subsequent dynamic LLM layer modeling. 

\begin{figure}[t]
\includegraphics[width=0.48\textwidth]{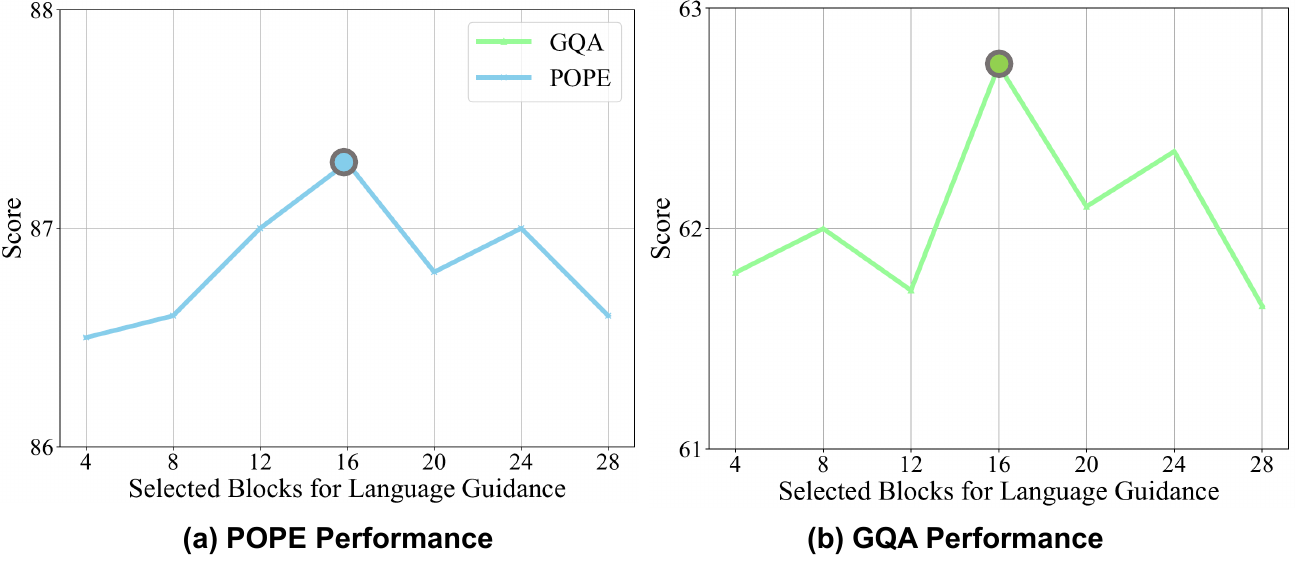}
\centering\caption{\textbf{Selected blocks for language guidance.
}}
\label{fig:language_guidance}
\end{figure}
\noindent\textbf{Analysis on the Inserted Blocks for Language Guidance.} We investigate the impact of inserting language guidance into different layers
of LLMs. 
We report the evaluation score of GQA and POPE datasets in Figure 4. We observe that the performance is low when we insert language guidance too early (\emph{i.e.}, 4, 8) as the model might not have gathered sufficient contextual information to generate effective guidance. Meanwhile, inserting language guidance too late (\emph{i.e.}, 24, 28) degenerates
the performance. We speculate this is due to the generated guidance being too concentrated and there not being enough layers to integrate the language-aware details.

\noindent\textbf{Analysis of Expert's Structure. } We systematically present the explicit benefits from the carefully designed expert's structure in Table~\ref{tab:expert}. The adapter-based structure surpasses MLP-based structure across all datasets, mainly due to the generated MLP is no longer a lightweight network to optimize, producing unstable performance. Compared with HyperNetwork+Adapter (Row 3 \emph{vs} Row 4), our proposed vision-language guided expert structure obtained the best performance. The results correspond with our assumption of the original HyperNetworks, which lacks a strong correlation between input and parameter generation. Our method, allows the model to make use of similarities between samples across datasets
and avoid potential interference between samples
within the same dataset.

\begin{figure}[t]
\includegraphics[width=0.48\textwidth]{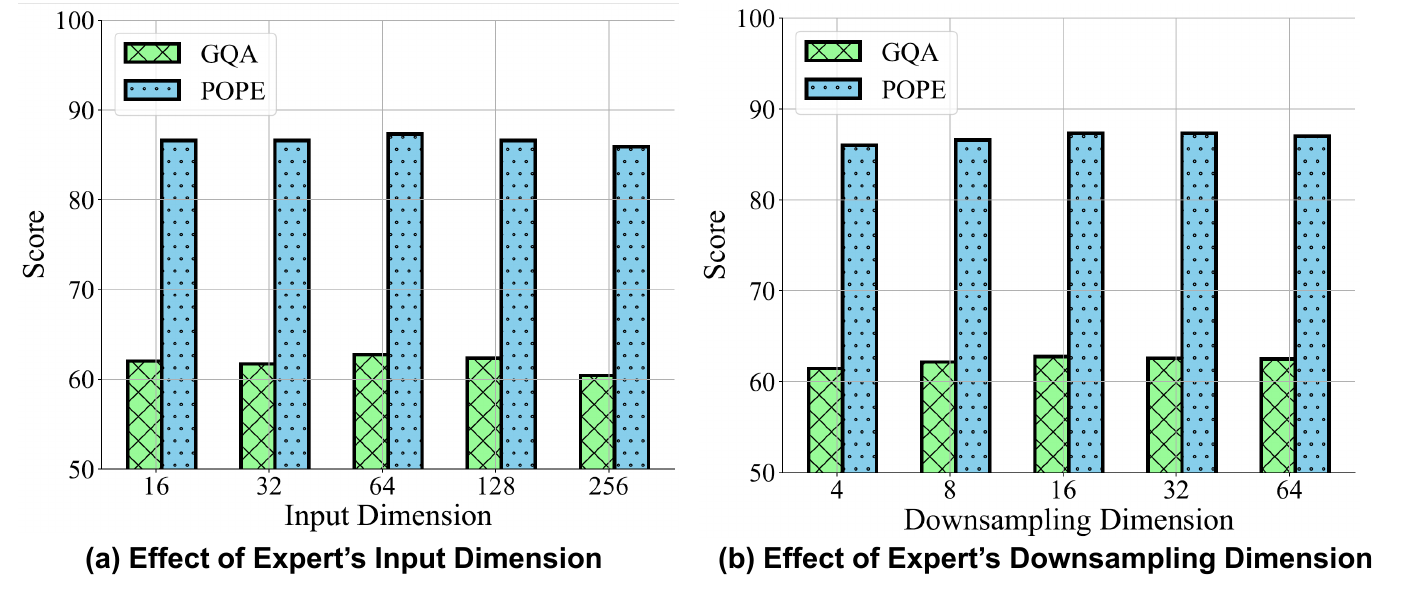}
\centering\caption{\textbf{Performance with respect to the different input and downsampling dimension in expert.}}
\label{fig:dimension}
\end{figure}

\noindent\textbf{Effect of Dimension of Expert Input and Downsampling.} Figure~\ref{fig:dimension} empirically provides an appropriate dimension of input and downsampling, \emph{i.e}, 64 and 16, respectively, either increasing or decreasing this value results in a performance decay. According to our analysis,  a bigger dimension may result in an unstable HyperNetwork optimization and a smaller value contains less language-guided information for dynamic learning, and thus yielding performance decay.

\noindent\textbf{Parameter-efficient Fine-tuning.} Our proposed language expert also can serve as a parameter-efficient fine-tuning function. The structure is similar to the HyperNetwork+Adapter. 
However, original hypernetwork-based approaches generally condition their parameters on a learned latent embedding, implying the model is the same for every example, yield performance decay. Summing up, the proposed language expert is an effective and parameter-efficient way to share information across multiple adapters to enable
positive transfer to low-resource and related tasks.

\noindent\textbf{Object Hallucination Evaluation.} 
We adopt the evaluation pipeline of POPE ~\cite{ref:POPE}, a polling-based query method, to evaluate object hallucination in \method{}. The results are presented in Table ~\ref{tab:object_hallucination},
where \method{} exhibits the best performance, indicating that \method{} tends to generate objects consistent with the given image. Additionally,
we observe that the ``yes'' ratio of \method{} remains relatively balanced, indicating that our model is capable of providing accurate feedback based on the questions.

\section{Conclusion}
Building upon HyperLLaVA's innovative dynamic tuning strategy, our work paves the way for groundbreaking advancements in multimodal learning systems. By adaptively tuning both projector and LLM parameters, and integrating dynamical visual and language experts, we not only surpass the performance benchmarks set by LLaVA but also introduce a parameter-efficient methodology. This approach offers a new horizon for enhancing multimodal task performances through personalized, dynamic adjustments. Future research could further explore the scalability of dynamic tuning mechanisms, potentially unlocking new avenues for understanding and integrating multimodal information more seamlessly.
 \clearpage

\bibliography{custom}

\appendix

\end{document}


\maketitle
 \clearpage
This is the Appendix for ``Revisiting the Domain Shift and Sample Uncertainty in Multi-source Active Domain Transfer''. Table~\ref{tab:abb} summarizes the abbreviations and the symbols used in the main body of our paper. 

\begin{table}[h]
\begin{center}
\captionsetup{font={small,stretch=1.25}, labelfont={bf}}
\caption{ Abbreviations and symbols used in the main paper.}
 \renewcommand{\arraystretch}{1.2}
\resizebox{0.7\textwidth}{!}
 {
  \begin{tabular}{c||c }
   \toprule[1.5pt]
   \textbf{Abbreviation/Symbol} & \textbf{Meaning}\\
   \hline
   \hline
   & \underline{\emph{Abbreviation}}\\
     Domain Adaptation&DA\\
      Unsupervised Domain Adaptation&UDA\\
      Active Domain Adaptation&ADA\\
   Multi-source Domain Adaptation &  MSDA\\
    Multi-source Active Domain Adaptation &  MADA\\
    Source Domain &  $\mathcal{D}_s$\\
    Multi-source  Domain &  $\{\mathcal{D}_{s,i}\}^M_{i=1}$\\
    Target Domain &  $\mathcal{D}_t$\\
     \hline
   \hline
     & \underline{\emph{Symbol in Algorithm}}\\
       Detective  & Dynamic integrated uncertainty valuation framework\\
       $\mathcal{M}(\cdot;(\Theta_{{m}},\Theta_{{a}}))$  & MADA Model\\
       $\mathcal{M}(\cdot;\Theta_{{m}})$  & Multi-domain Adaptation Model\\
       $\mathcal{M}(\cdot;\Theta_{{a}})$  & Active Learning Model\\
       $Dir(\cdot)$  & Dirichlet Distribution\\
       $H(\cdot)$  & Shannon Entropy\\
       $\mathcal{U}_{dom}$  & Domain Uncertainty \\
       $\mathcal{U}_{pre}$  & Predictive Uncertainty\\
       $\mathcal{U}_{ins}$  & Integrated Uncertainty\\
       $\mathcal{D}_{con}$  & Contextual Diversity\\
       UDN  & Universal Dynamic Network\\
      IUS  & Integrated Uncertainty Selector\\
        CDC  & Contextual Diversity Calculator\\
        $\mathcal{L}_{mar}$  & Marginal Likelihood  Loss\\
        $\mathcal{L}_{kl}$  & KL-divergence Loss\\
   \toprule[1.5pt]
  \end{tabular}
  }
  \label{tab:abb}
\end{center}
\end{table}

This appendix is organized as follows:

\begin{itemize}

\item Section ~\ref{sec:1} provides the method details of proposed \method{}, including Integrated Uncertainty Selector (IUS) and Contextual Diversity Calculator (CDC).

\item Section ~\ref{sec:2} reports more experimental settings of baselines, implementation details and training process of \method{}.

\item Section ~\ref{sec:3} shows the additional experiments on the \texttt{Office-Home}~\cite{venkateswara2017deep} and \texttt{Digits-five}~\cite{nguyen2021most} dataset to further verify the effectiveness of \method{}.

\item Section ~\ref{sec:4} lists the limitations of this paper.
\end{itemize}

\subsection{HyperLLaVA Training}
We optimize the output of \method{} through a generative loss in an auto-regressive manner. Given an image and text, \method{} generates the output sequence $R = [r_1, r_2, \cdots, r_k]$ by progressively
generating each element, where K = P + D represents the
length of the output sequence. The formula is:
\begin{align}
\label{eq:adapter}
\begin{gathered}
L=-\sum_{i=1}^K(R^{P+i}|V,T^{[:i-1]})
\end{gathered}
\end{align}

\noindent\textbf{Implementation Details.}  
In the training of the \method{}, we utilize the ADAMW optimizer, adapting hyperparameters to cater to the specific requirements of each phase. For the feature alignment stage, parameters are set as \(B=32\), \(Lr=0.001\), while for the visual instruction tuning stage, we adjust the parameters to \(B=16\), \(Lr=0.00002\). The configuration for the ADAMW optimizer incorporates the following settings: \(\boldsymbol{\beta} = (0.9, 0.999)\), \(\varepsilon = 1 \times 10^{-8}\), and $W_d$ = 0.0, ensuring a bespoke optimization strategy that effectively addresses the unique demands of each training phase.

We train our model following the same training process as LLaVA-1.5. The process includes two stages: (1) feature alignment stage: use 558K subset of the LAION-CC-SBU dataset to connect a frozen pretrained vision encoder to a frozen LLM; (2) visual instruction tuning stage: use 150K GPT-generated multimodal instruction-following data, plus around 515K VQA data from academic-oriented tasks, to teach the model to follow multimodal instructions.

\bibliography{custom}

\appendix